\pdfoutput=1

\documentclass[11pt]{article}

\usepackage[final]{acl}

\usepackage{times}
\usepackage{latexsym}

\usepackage[T1]{fontenc}

\usepackage[utf8]{inputenc}

\usepackage{microtype}

\usepackage{inconsolata}
\usepackage{enumitem}

\usepackage{xcolor}

\usepackage{graphicx}
\usepackage{url}
\usepackage{booktabs}

\title{Error-preserving Automatic Speech Recognition of Young English Learners' Language}

\author{
Janick Michot\textsuperscript{1}, Manuela Hürlimann\textsuperscript{1}, Jan Deriu\textsuperscript{1}, Luzia Sauer\textsuperscript{2}.  \\ 
{\bf Katsiaryna Mlynchyk \textsuperscript{1}, }{\bf Mark Cieliebak\textsuperscript{1}} \\
\textsuperscript{1} Zurich University of Applied Sciences, Winterthur \\
\textsuperscript{2} Pädagogische Hochschule Zurich, Zurich\\
mict@zhaw.ch, hueu@zhaw.ch, deri@zhaw.ch, luzia.sauer@phzh.ch, \\
mlyn@zhaw.ch, ciel@zhaw.ch
}

\begin{document}
\maketitle
\begin{abstract}
One of the central skills that language learners need to practice is speaking the language. Currently, students in school do not get enough speaking opportunities and lack conversational practice. Recent advances in speech technology and natural language processing allow for the creation of novel tools to practice their speaking skills. In this work, we tackle the first component of such a pipeline, namely, the automated speech recognition module (ASR), which faces a number of challenges: first, state-of-the-art ASR models are often trained on adult read-aloud data by native speakers and do not transfer well to young language learners' speech. Second, most ASR systems contain a powerful language model, which smooths out errors made by the speakers. To give corrective feedback, which is a crucial part of language learning, the ASR systems in our setting need to preserve the errors made by the language learners. In this work, we build an ASR system that satisfies these requirements: it works on spontaneous speech by young language learners and preserves their errors. For this, we collected a corpus containing around 85 hours of English audio spoken by learners in Switzerland from grades 4 to 6 on different language learning tasks, which we used to train an ASR model. Our experiments show that our model benefits from direct fine-tuning on children's voices and has a much higher error preservation rate than other models. 
\end{abstract}



\section{Introduction}

Speaking is one of the core competencies to be developed in foreign language classes and 
the second most widely used skill in everyday-life communication~\cite{hedge2001teaching}. For students to successfully acquire speaking competencies, they must be trained from an early stage in the language learning process and in a systematic manner. However, speech production is a highly complex process that is often not addressed adequately in classrooms. The main issue is that students often do not get enough speaking opportunities~\cite{kleinschroth2014sprechkompetenz,grimm2015teaching}, and lack extended conversational practice~\cite{Pfenninger_Lendl_2017}. The recent advancements in both speech processing~\cite{malik2021automatic}, and conversational dialogue systems~\cite{deriu2021survey,ni2023recent} provide an opportunity to increase the speaking practice of language learners using automated tools.

The work presented in this paper is part of a larger effort to develop an interactive, voice-driven chatbot with which learners can practice their interactive speaking skills. The bot is designed as a conversation partner that adjusts to the proficiency level and interests of the students and provides corrective feedback to support their language development.

One key issue is the automated speech recognition (ASR) module, which transcribes the utterances of the language learners into text to be processed in downstream tasks (e.g., speaker-error analysis, dialogue systems, inter alia). The focus of this work is to adapt the ASR module to handle children's speech in a language learning environment. The core challenge for the ASR system in this setting is not only to transcribe the speech but to make sure that the errors made by the language learners are transcribed faithfully. This is needed to provide language learners with corrective feedback, which is a key component of foreign language development. It prompts learners to notice errors and is likely to lead to utterance repair, which, in turn, facilitates language development \cite{ellis2021explicit}. Our investigations showed that current state-of-the-art ASR models tend to correct the speakers' errors, which renders giving corrective feedback impossible. 

The second challenge for the ASR system is handling spontaneous children's speech since most of these systems are trained on adult read-aloud error-free corpora recorded by native speakers~\cite{panayotov2015librispeech, commonvoice:2020}. Children's speech, especially spontaneous speech of language learners, differs significantly from read-aloud speech of native adult speakers~\cite{gurunath2020trafo}. Children's' speech has a different range of sound frequencies~\cite{potamianos2003robust_child}, a high within-subjects variability~\cite{gerosa2006acoustic} and a high inter-speaker variability in different age groups~\cite{lee1999acoustics}. 

These challenges yield three research questions, which we address in this work:
\begin{enumerate}[noitemsep,topsep=0pt]
    \item How can we measure error preservation, i.e. the "verbatimness" of an ASR transcript?
    \item How well do current pre-trained ASR systems perform on learners’ spontaneous English productions, with respect to error preservation and in general?
    \item Does fine-tuning pre-trained systems with data from young learners lead to improved error preservation in the ASR transcripts?
\end{enumerate}


\paragraph{Contributions} In order to answer these questions, we first collected a dataset of young learners in Swiss public schools speaking English, consisting of 85 hours of recordings corresponding to 45'004 individual utterances by 327 distinct speakers. We subsequently created verbatim transcriptions of these recordings, in which learner errors are annotated using specific symbols. This dataset can be accessed on HuggingFace\footnote{\url{https://huggingface.co/datasets/mict-zhaw/chall}}, but the dataset files must be downloaded manually as described in Section \ref{subsec:availability} below. We next developed a metric for error preservation, called \emph{Word-Based Error Preservation Rate (WEPR)}, which takes into account only those reference words that contain an error annotation. Using WEPR and standard ASR metrics, we compared 7 pre-trained ASR systems with a custom fine-tuned model\footnote{\url{https://huggingface.co/mict-zhaw/chall\_wav2vec2\_xlsr\_300m}\label{model_footnote}}. Our results show that a) there are large differences between the pre-trained models both in terms of standard metrics and in terms of WEPR and b) fine-tuning significantly improves error preservation of learners' speech. All related code can be accessed on GitHub\footnote{\url{https://github.com/mict-zhaw/chall\_e2e\_stt}}.


\section{Related Work}

\noindent\textbf{Children's Speech Corpora.} Corpora of children's speech can be divided into two types: i) corpora for native speaking children intended for building virtual tutors for non-language subjects, ii) corpora for young language learners that support building virtual tutors for language learning. 

The MyST Children's Speech Corpus~\cite{pradhan-etal-2016-science,ward2019my} contains 499 hours of conversational speech  (out of which 233 hours are manually transcribed) for a virtual tutor for science topics targeted at young English native speakers. The OGI Kids' Speech Corpus~\cite{shobaki2000ogi} contains spontaneous speech from 1100 American children from kindergarten through grade 10, mainly consisting of scripted speech in the form of words and utterances, and a small sample of spontaneous speech. The AusKidTalk corpus~\cite{ahmed21auskidstalk} contains speech from Australian children ages 3 to 12 consisting of single words, utterances, and narrative speech. Other, smaller, datasets of native speaking children are available for different purposes such as read-aloud support~\cite{eskenazi1996kids} or general analysis of English children's speech~\cite{lee1999acoustics,hagen2003children}. For German, the KidsTalk corpus~\cite{rumberg2022kidsTalk} contains 25 hours of transcribed continuous speech from children aged 3 to 11. All these corpora are devised for settings with native speakers. 

For language learners, there are far fewer datasets of children's speech. The TLT-school collection~\cite{gretter2020tlt} aims at assessing the proficiency of 9- to 16-year old Italian native speakers in English and German. TLT was recorded with a pool of 3000 students, resulting in approximately 275h of English and 265h of German data, out of which 16h for English and 8h for German have been transcribed. The corpus closest to our dataset is the CALL corpus~\cite{baur2018call_task}, consisting of English utterances by Swiss German second and third year learners, where the task is to label the correctness of each utterance. In total, the corpus contains 38k utterances of students interacting with an online dialogue system, where they receive various prompts to produce speech. Across a series of shared tasks, subsets of around 6k annotated utterances have been released. The setting differs significantly from ours as we are interested in spontaneous speech with transcriptions to train an ASR system which can automatically transcribe learners' speech verbatim.  \citet{batliner05b_interspeech} introduce a children's speech corpus containing 60 hours of children's speech aged 4-13 in a variety of languages, such as English, German, and Swedish, as well as English speech from German, Italian, and Swedish children. In general, there is only a limited amount of work investigating the effects of fine-tuning models on speech data of language learners to retain the speakers' errors. ~\citet{ma2023slate} fine-tune Whisper to investigate its ability to retain hesitations, numbers, abbreviations, disfluencies, and incomplete words.  Instead, we aim to preserve speaker errors in grammar, lexical choice, and pronunciation.

\noindent\textbf{ASR for children's speech and language learners.} 
The literature on ASR models for children's speech, especially for non-native language learners, is sparse. Most notably, \citet{lu2022improving} investigated the performance of fine-tuning wav2vec 2.0~\cite{baevski2020wav2vec} on children's speech (both native MyST and OGI), as well as non-native speech (TLT) compared to fine-tuning on adult-only data. The results show that ASR models trained on children's speech significantly outperform those models trained on adult-speech only, even in the case of non-native speakers. Similarly, \citet{gurunath2022e2e_stt4child} investigated the impact of using children's data for fine-tuning ASR models. The conclusion is similar to \citet{lu2022improving}: adding children's data yields better performance; however, the performance of an adult ASR model on adult data is higher than the performance of an ASR model trained and applied on children's data. While both \citet{lu2022improving} and \citet{gurunath2022e2e_stt4child} are interested in the overall performance in terms of WER, our work focuses on the preservation of errors made by non-native children. 


\section{Dataset: Spontaneous Speech of Young Learners of English}

We now describe the dataset that we collected for the purpose of this research. It contains 85 hours of audio recordings of spontaneous speech by young young learners of English in Switzerland. Each recording is paired with a verbatim transcript that contains error annotations. 

\subsection{Audio Recording}\label{subsec:audio-recording}
The recording setup was designed such that the collected speech resembled the kind of conversations intended for the learners to hold with the chatbot.
We used playful and engaging activities targeted to elicit extended authentic communication from young learners. Activities included role plays with problem-solving components (e.g. ‘going shopping for a school trip’), guessing games (e.g. riddles), TV interviews with imaginary characters and asking/answering personal questions (e.g. ‘if you could go into space, what would you take with you?’). All activities were piloted with a grade 4 class and maintained, adjusted (to yield more data) or rejected (e.g. because the task led to students communicating non-verbally and/or with much noise) for the main data collection period.  To support learners, each activity further included visual and language support (e.g. cartoon characters they could choose from, sample dialogues, language chunks) as well as a preparation phase during which the students could familiarise themselves with the tasks by use of example sentences and model dialogues.\footnote{The descriptions of the speaking activities is provided in the repository.}

\noindent\textbf{Speaker recruitment and consent}
After receiving permission to collect audio data with minors from key government institutions that act as ethics review boards in Switzerland concerning research with schools and their learners, we recruited 20 primary school teachers interested in participating in our project with their classes (via personal and university networks, newsletters and direct contact with schools). Participation was entirely voluntary and could be withdrawn at any time. Participation further necessitated the approval of the school principal and the written consent of each student’s legal caretaker.\footnote{We share the consent forms in our repository.}

In the span of 9 months (March-November 2023), 337 primary school students aged 9 to 14 years (4th to 6th graders) enrolled in 8 different schools in German-speaking Switzerland performed our activities in pairs, trios or alone (if necessary) in three different settings: at school recorded by project members; on the university campus recorded by project members and student assistants; and at school recorded by teachers and sent to us via safe weblinks.  For reasons of practicability/feasibility (i.e., to respect teachers' tight schedules, time, and finances), the corpus was not annotated for CEFR levels, but according to the Swiss curriculum LP21, it should reflect performance at the A1 and A2 levels (English Basic Users). Some participants, including native-speaking children, performed beyond these levels. 
School principals, teachers, and students were not remunerated for their participation but received small tokens of appreciation, such as flowers and chocolates.

\noindent\textbf{Metadata} Each recording is associated with the following metadata:
\begin{itemize}
    \itemsep0.1em 
    \item School area code: an integer between 1 and 8 (inclusive)
    \item School grade of the speakers: 4gr, 5gr, 6gr as well as combinations (4/5gr, 5/6gr, 4/6gr)
    \item Recording Device
    \item Recording Application
    \item Speaking activities
    \item Background Noise: a boolean indicating whether background noise is audible in the recording (set manually by project members).
\end{itemize}

\subsection{Transcription and Error Annotation}
\label{subsec:transcription-annotation}


The transcription of our voice data was outsourced to a transcription agency. Services included both the transcription of the voice data and the annotation of lexical, grammatical and pronunciation errors, as well as usage of German words. We developed a comprehensive data transcription guideline for the transcription agency which was first piloted on a small number of transcripts and then adjusted where necessary. 
Transcription guidelines included information about spelling conventions (British English), the frequency and nature of timestamps (start and end time of each word, in milliseconds), error codes (@! for errors of any kind and @g for German words) and disfluency markers (e.g. a hyphen "–" for verbatim repetitions, such as ‘he's -- he's really tall’). The complete transcription guidelines are provided in the supplementary material of this paper.

\subsection{Data Aggregation and Filtering}\label{subsec:data-agg-cleaning}
The recording stage resulted in 1039 audio recordings. Of these, 23 were removed due to missing metadata or missing/retracted consent, so a total of 1016 recordings and their associated metadata and transcriptions were available for our experiments.

These recordings were split into individual utterances by a single speaker using the word-level timestamps provided in the transcripts, resulting in 49'608 utterances.We removed utterances shorter than 0.5 seconds and utterances attributed to adults (e.g. short interventions by teachers), creating a final dataset of 45'004 utterances corresponding to 85 hours of audio.  Each utterance was paired with its reference transcription and metadata.

\subsection{Final Dataset}

\begin{figure*}[t!]
\centering
\includegraphics[width=0.75\textwidth]{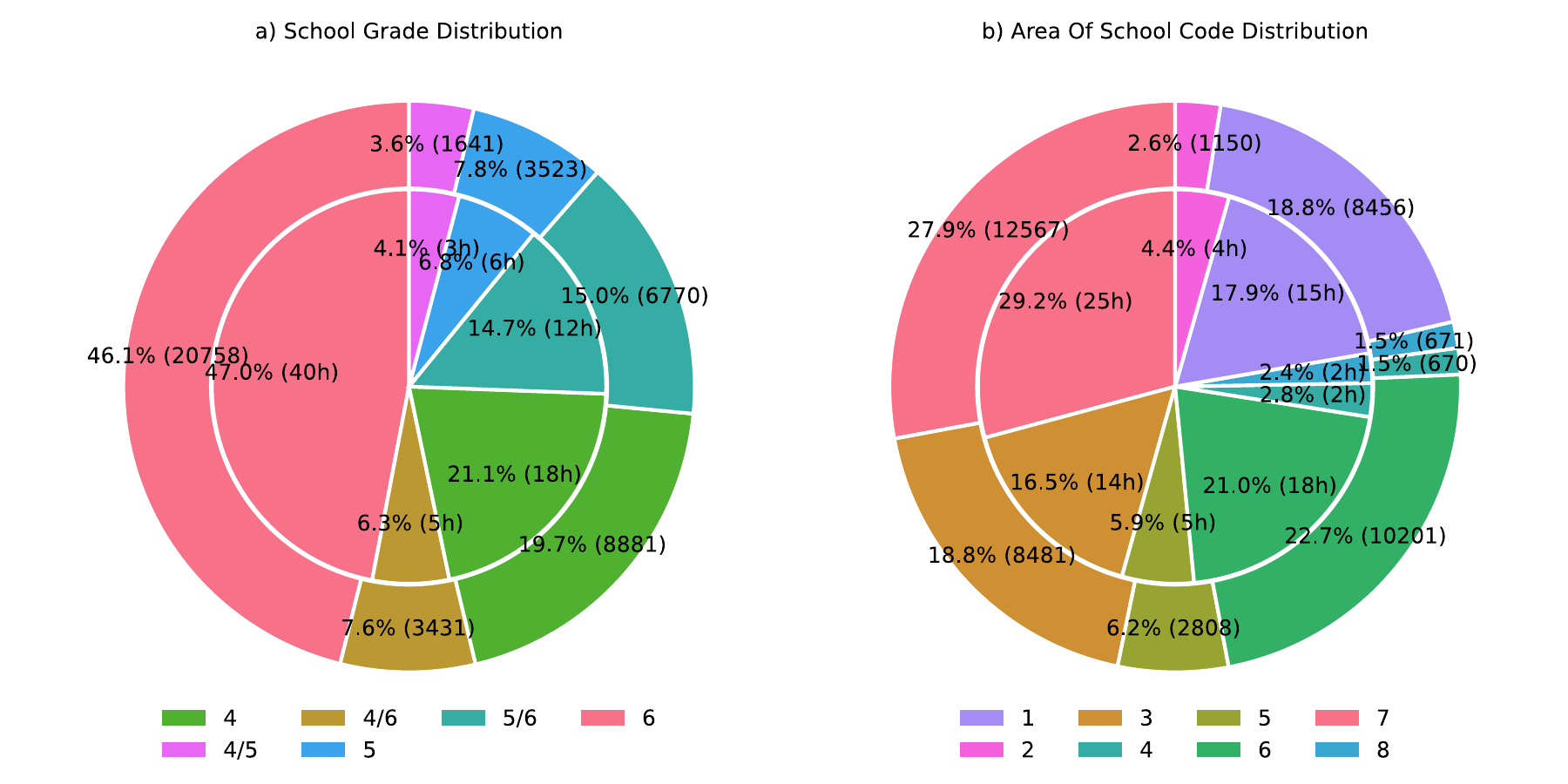}
\caption{Number of utterances (outer ring) and audio hours (inner ring) by school grade (a) and school area code (b).}
\label{fig:utterances-hours-grade-area}
\end{figure*}

The final dataset contains 45'004 utterances by 327 distinct speakers. Figure \ref{fig:utterances-hours-grade-area} shows the number of recordings and audio duration by school grades and school area codes. Almost half the data in terms of both utterances and hours comes from 6th graders, while the other half is split among the other grades.
The dataset contains 485,770 tokens and 10,203 distinct types. There are 14,396 error-annotated tokens with 2,004 underlying types. Thus, our data contains a large amount of tokens and a relatively large amount of token diversity.

\begin{figure}[t!]
\small
\centering
\includegraphics[width=0.47\textwidth]{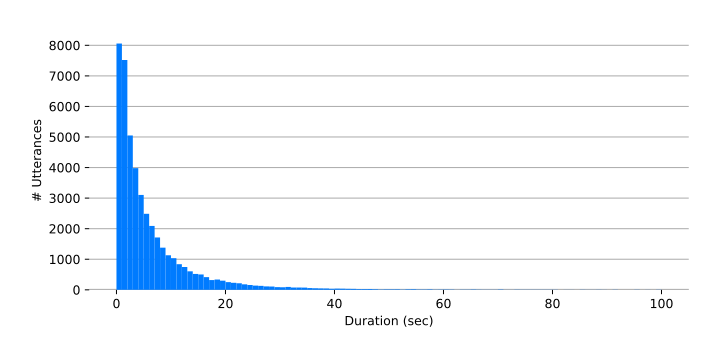}
\caption{Distribution of utterance lengths.}
\label{fig:utterance-lenght-dist}
\end{figure}

The length distribution is shown in Figure \ref{fig:utterance-lenght-dist}. It can be seen that most utterances are between 0.5 and 20 seconds long.

\subsubsection{Data Folds} \label{subsubsec:datafolds}

\begin{figure}[t!]
\small
\centering
\includegraphics[width=0.45\textwidth]{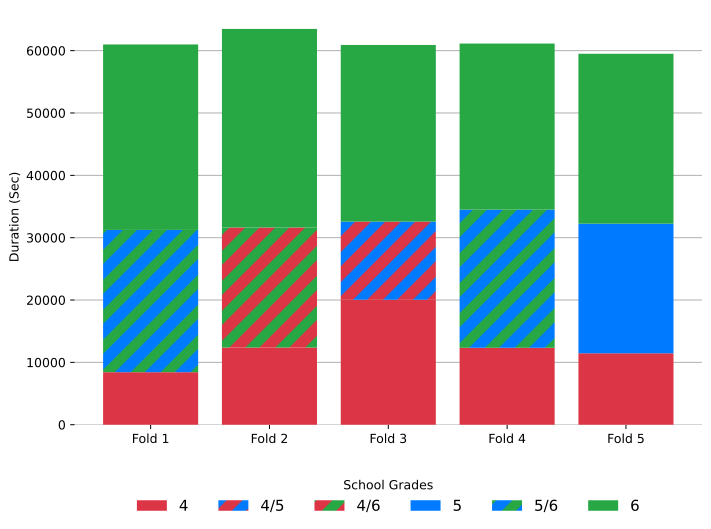}
\caption{Duration and grade distribution of the data folds.}
\label{fig:data-folds}
\end{figure}

For the experiments in this paper, we split the dataset into five distinct folds of similar duration (about 16h each), where each class (and therefore also each speaker) occurs in only one fold. To simulate the use case of the ASR system being confronted with a new class of learners, each fold contains data from a mix of grades. Figure \ref{fig:data-folds} visualises the duration and grade distribution of each fold.

\subsubsection{Data Availability}
\label{subsec:availability}
The dataset that we collected contains sensitive data of minors and thus cannot be shared publicly. The data can, however, be accessed as part of a joint project with one or several of the original project partners, subject to a collaboration agreement. Before sharing, all transcripts will undergo complete anonymisation so that any names and other personal information are removed.

\section{Error-Preserving Automatic Speech Recognition}

This section presents the metrics used for measuring error preservation and evaluating systems (Section \ref{subsec:metrics}), as well as the approaches to comparing pre-trained ASR systems (Section \ref{subsec:pre-trained}) and to fine-tuning existing systems using our learner dataset (Section \ref{subsec:finetuning}). The qualitative results are presented and discussed in Section \ref{subsec:quant-results}, and a qualitative evaluation is shared in \ref{subsec:qual-results}.

\subsection{Metrics}\label{subsec:metrics}

In order to measure error preservation, we use the error annotations that were manually added to each utterance (cp. Section \ref{subsec:transcription-annotation}) and a custom phonetic word-level alignment algorithm. This algorithm aligns two or more sequences (e.g., a reference and one or multiple hypotheses), identifying matches, substitutions (S), insertions (I), and deletions (D) at the word level. Our metric, WEPR (Word-Based Error Preservation Rate), considers only those word pairs where the reference word contains an error annotation. WEPR is calculated according to equation \ref{eq:wepr}:  
$\mathcal{A}$ is the set of annotations that are considered (e.g. $\mathcal{A = \{\mathtt{@!}, \mathtt{@g}\}}$), $\mathcal{S}$ and $\mathcal{D}$ are the number of substitutions and deletions, respectively, where the reference word contains an error annotation, and $\mathcal{N}$ is the total number of reference words that contain an error annotation.

\begin{equation}
\small
\textrm{$WEPR(\mathcal{A})=  \frac{(\mathcal{S+D})}{\mathcal{N}}$}
\label{eq:wepr}
\end{equation}

In addition to WEPR, we also compute the following general ASR metrics using all words in the utterance: Word Error Rate (WER)\footnote{\url{https://github.com/huggingface/evaluate/blob/main/metrics/wer/wer.py}}, Character Error Rate (CER)\footnote{\url{https://github.com/huggingface/evaluate/blob/main/metrics/cer/cer.py}}, and character n-gram F-Score (chrF)\footnote{\url{https://www.nltk.org/api/nltk.translate.chrf_score.html\#nltk.translate.chrf_score.corpus_chrf}} \cite{popovic2015chrf}.

We evaluate all models on our dataset's five folds (cp. Section \ref{subsubsec:datafolds}) and report for each model the mean and standard deviation across all folds.

For evaluation, all texts are normalised using the Whisper normalizer for English \footnote{\url{https://github.com/openai/whisper/blob/main/whisper/normalizers/english.py}}. Normalizing texts can mitigate the impact of disfluencies and non-standard linguistic forms, common in non-native and children's speech. This allows for a more accurate comparison between different ASR models, as it aligns the hypothesis and reference texts more closely.

Table~\ref{tab:wepr} demonstrates text normalization and WEPR calculation using a Whisper Large model prediction and a modified normalizer. The adjustments to the Whisper normalizer maintain contractions in references, ensuring the integrity of annotated words (e.g., it’s@!). For error preservation assessment, phonetic alignment is performed between the normalized prediction (NP) and a reference without error annotations (NR2). This removal is essential because the alignment algorithm would otherwise misidentify annotated words. The normalized reference with annotations (NR1) is then used to identify word pairs in the alignment results that are relevant for WEPR calculation. The Whisper Normalizer’s tendency to increase the similarity between predictiosn and references helps in building and classifying these word-level pairs.

\begin{table*}[ht!]
\centering
\resizebox{.95\textwidth}{!} {
    \begin{tabular}{|p{1cm} p{5cm}|p{15cm}|}
    \hline
    & Category & Content \\
    \hline
    R & Reference & The beach, because it's \textbf{a@!} very nice \textbf{of@!} the beach. Tell me about \textbf{you@!} favorite TV-show. \\
    P & Prediction & the beach because it's a very nice beach tell me about your favorite TV show \\
    \hline
    NR1 & Normalized Reference (with @!) &  the beach because it's \textbf{a@!} very nice \textbf{of@!} the beach tell me about \textbf{you@!} favorite tvshow \\
    NR2 & Normalized Reference & the beach because it's a very nice of the beach tell me about you favorite tvshow \\
    NP & Normalized Prediction & the beach because it's a very nice beach tell me about your favorite tv show \\
    \hline
    S & Substitutions & [(\textbf{'you@!'}, 'your')] \\
    D & Deletions & [\textbf{'of@!'}] \\
    I & Insertions & [] \\
    C & Correct & [\textbf{'a@!'}] \\
    \hline
    & WEPR & \textbf{0.67} \\
    \hline
    \end{tabular}
}
\caption{WEPR calculation example using Whisper Large model's prediction with texts normalized by a customized version of Whisper’s Text Normalizer retaining contractions. Substitutions, insertions, deletions, and correct words are derived from phonetic alignment between NR2 and NP, but only for words that are annotated in NR1.}
\label{tab:wepr}
\end{table*}

\begin{table*}[ht!]
    \centering
    \resizebox{.75\textwidth}{!}{%
            \begin{tabular}{|l|c|c|c|c|c|}
    \hline
    System Name& \#Param.& WER & CER & chrF & WEPR\\ \hline
    Wav2Vec Base                      &95M&$0.55\pm0.02$&$0.34\pm0.02$&$0.35\pm0.02$&$0.57\pm0.02$\\
    Wav2Vec Large                     &317M&$0.49\pm0.02$&$0.29\pm0.01$&$0.41\pm0.02$&$0.50\pm0.02$\\
    XLSR-53 + CommonVoice 6.1         &317M&$0.38\pm0.01$&$0.26\pm0.01$&$0.59\pm0.01$&$0.50\pm0.03$\\
    XLSR-1B + CommonVoice 6.1         &1B&$0.31\pm0.01$&$0.21\pm0.01$&$0.61\pm0.01$&$0.44\pm0.03$\\
    Whisper Medium                    &769M&$0.26\pm0.02$&$0.20\pm0.03$&$\textbf{0.70}\pm0.02$&$0.46\pm0.04$\\
    Whisper Large                     &1.5B&$\textbf{0.25}\pm0.02$&$0.19\pm0.01$&$\textbf{0.70}\pm0.01$&$0.47\pm0.03$\\
    Whisper Large-v3                  &1.5B&$0.30\pm0.04$&$0.23\pm0.03$&$\textbf{0.70}\pm0.02$&$0.45\pm0.03$\\ \hline
    ChaLL-300M (ours)                 &300M&$0.30\pm0.01$&$\textbf{0.16}\pm0.01$&$0.68\pm0.01$&$\textbf{0.38}\pm0.03$\\ \hline
    \end{tabular}%
        }
    \caption{Results of the 5-fold evaluation. We report for each model the mean and standard deviation (mean$\pm$std) of the scores on each of the 5 folds. The bottom row shows the scores of our fine-tuned model. }
    \label{tab:results}
\end{table*}

\subsection{Pre-trained ASR Systems}\label{subsec:pre-trained}
We compare the performance of state-of-the-art ASR systems trained on datasets of adult English speakers. For this, we select seven different models, four based on a CTC decoding strategy, and three based on an encoder-decoder architecture. Our hypothesis is that CTC models are better at preserving speaker-errors as they do not rely on a language model, which potentially corrects such errors. Therefore, we do not use a n-gram language model during the CTC decoding phase, which is usually added for better WER performance. For the CTC-based models, we use the original Wav2VWec 2.0 large and base models~\cite{baevski2020wav2vec} fine-tuned on 960h of Librispeech~\cite{panayotov2015librispeech} (English adult read-aloud data). We also use the fine-tuned Wav2Vec 2.0 models provided by~\citet{grosman2021xlsr53-large-english,grosman2021xlsr-1b-english}, which are based on the XLSR pretraining~\cite{babu2021xls}, and were fine-tuned on the CommonVoice 6.1 data~\cite{commonvoice:2020} consisting of approximately 2100 hours of English adult read-aloud data. For the encoder-decoder architecture, we used the Whisper medium, large, and large-v3 models provided by OpenAI~\cite{radford2022robust}.







\begin{table*}[t!]
    \centering
    \small
    \resizebox{.65\textwidth}{!}{%
     \begin{tabular}{|l|l|l|l|l|}
 \hline
  \textsc{Target}  & \textsc{Prediction} & \textsc{{Chall-300M}} & \textsc{Whisper-Large}  & \textsc{XLSR-1B} \\
    \hline
de@! & the & 0.946 & 0.869 & \textbf{ 0.805 } \\
\hline
a@! & \_ & \textbf{ 0.114 } & 0.327 & 0.347 \\
\hline
a@! & an & \textbf{ 0.026 } & 0.398 & 0.257 \\
\hline
have@! & has & \textbf{ 0.015 } & 0.231 & 0.052 \\
\hline
have@! & \_ & \textbf{ 0.034 } & 0.128 & 0.129 \\
\hline
you@! & your & 0.244 & 0.306 & \textbf{ 0.099 } \\
\hline
it's@! & it & \textbf{ 0.068 } & 0.116 & 0.136 \\
\hline
it's@! & \_ & \textbf{ 0.043 } & 0.119 & 0.146 \\
\hline
is@! & \_ & \textbf{ 0.05 } & 0.125 & 0.136 \\
\hline
it's@! & is & \textbf{ 0.055 } & 0.133 & 0.103 \\
\hline
are@! & \_ & \textbf{ 0.072 } & 0.162 & 0.144 \\
\hline
dis@! & this & 0.854 & 0.83 & \textbf{ 0.717 } \\
\hline
it@! & \_ & \textbf{ 0.094 } & 0.311 & 0.193 \\
\hline
he@! & \_ & \textbf{ 0.175 } & 0.254 & 0.356 \\
\hline
de@! & \_ & \textbf{ 0.029 } & 0.123 & 0.143 \\
\hline
the@! & \_ & \textbf{ 0.046 } & 0.24 & 0.183 \\
\hline
in@! & \_ & \textbf{ 0.027 } & 0.127 & 0.107 \\
\hline
you@! & \_ & \textbf{ 0.077 } & 0.113 & 0.117 \\
\hline
i@! & \_ & \textbf{ 0.133 } & 0.248 & 0.294 \\
\hline
on@! & \_ & \textbf{ 0.019 } & 0.129 & 0.105 \\
\hline
\hline
\multicolumn{2}{|c|}{Mean (n=20)} & \textbf{0.156} & 0.265 & 0.229 \\
\hline
\end{tabular}
    }
    \caption{System comparison on 20 most frequent incorrectly transcribed speaker-errors. For each system, the number indicates the fraction of cases in which the system incorrectly transcribes the error \textsc{Target} as \textsc{Prediction} (where "\_" denotes deletion of \textsc{Target}). The lowest value of each row is set in boldface. The final row shows the mean across the 20 samples.}
    \label{tab:confusion_ratio}
\end{table*}

\begin{table*}[t!]
    \centering
    \small
    \resizebox{.65\textwidth}{!}{%
     \begin{tabular}{|l|l|l|l|l|}
 \hline
  \textsc{Target}  & \textsc{Prediction} & \textsc{{Chall-300M}} & \textsc{Whisper-Large}  & \textsc{XLSR-1B} \\
    \hline
have@! & have & \textbf{ 0.875 } & 0.587 & 0.699 \\
\hline
a@! & a & \textbf{ 0.804 } & 0.215 & 0.325 \\
\hline
is@! & is & \textbf{ 0.79 } & 0.667 & 0.769 \\
\hline
in@! & in & \textbf{ 0.897 } & 0.773 & 0.807 \\
\hline
it's@! & it's & \textbf{ 0.703 } & 0.568 & 0.482 \\
\hline
are@! & are & \textbf{ 0.739 } & 0.688 & 0.699 \\
\hline
on@! & on & \textbf{ 0.917 } & 0.749 & 0.79 \\
\hline
of@! & of & \textbf{ 0.922 } & 0.705 & 0.848 \\
\hline
the@! & the & \textbf{ 0.815 } & 0.632 & 0.678 \\
\hline
you@! & you & 0.606 & 0.55 & \textbf{ 0.735 } \\
\hline
she@! & she & \textbf{ 0.867 } & 0.713 & 0.774 \\
\hline
it@! & it & \textbf{ 0.772 } & 0.579 & 0.659 \\
\hline
has@! & has & \textbf{ 0.825 } & 0.775 & 0.774 \\
\hline
make@! & make & \textbf{ 0.95 } & 0.746 & 0.808 \\
\hline
do@! & do & \textbf{ 0.82 } & 0.744 & 0.748 \\
\hline
much@! & much & \textbf{ 0.98 } & 0.96 & 0.96 \\
\hline
he@! & he & \textbf{ 0.679 } & 0.627 & 0.561 \\
\hline
not@! & not & \textbf{ 0.89 } & 0.75 & 0.777 \\
\hline
at@! & at & \textbf{ 0.811 } & 0.612 & 0.759 \\
\hline
don't@! & don't & \textbf{ 0.885 } & 0.826 & 0.811 \\
\hline
\hline
\multicolumn{2}{|c|}{Mean (n=20)} & \textbf{0.827} & 0.673 & 0.723 \\
\hline
\end{tabular}
    }
    \caption{System comparison on 20 most frequent correctly preserved speaker-errors. For each system, the number indicates the fraction of cases in which the system correctly transcribes the error \textsc{Target} as \textsc{Prediction}. The highest value of each row is set in boldface. The final row shows the mean across the 20 samples.}
    \label{tab:correct_ratio}
\end{table*}

\subsection{Fine-tuning Pre-trained ASR Systems Using Learner Data}\label{subsec:finetuning}
To evaluate the impact of fine-tuning, we fine-tune the Wav2Vec-XLSR-300M model~\footnote{Due to the high computational cost, we decided to use the 300M model instead of the 1B model.}~\cite{babu2021xls} on our collected language learner data.

\noindent\textbf{Data Preprocessing.}
For fine-tuning, we split longer utterances into chunks of a maximum of 12 seconds and removed trailing pauses. The transcripts were preprocessed as follows:
\begin{itemize}[noitemsep]
    \itemsep0.1em 
    \item Remove error annotations and other transcript conventions 
    \item Convert to lowercase
    \item Standardise text (Remove text between brackets and parentheses. Standardise apostrophes by removing spaces before them. Remove commas between digits and periods not followed by numbers.)
    \item Clean and standardise whitespace
    \item Normalise/remove special characters.
    \item Transform numbers into words using \textit{num2words}
\end{itemize}

\noindent\textbf{Approach.} We apply 5-fold cross-validation (cf. \ref{subsubsec:datafolds}), that is, we train on four folds, and evaluate on the held-out fold. We trained each run on 6 nVidia Tesla V100 GPUs for 4000 steps using a learning rate of 3e-5, a per-device batch size of 14, and 15 gradient accumulation steps (for a total batch size of 1260, which corresponds to approx. 2 hours of audio per batch), and we used the 8-bit AdamW optimizer~\cite{loshchilov2017decoupled,dettmers20218}. 
Our fine-tuned model, called ChaLL-300M, is available on HuggingFace.\footref{model_footnote}



\subsection{Quantitative Results}\label{subsec:quant-results}

\noindent\textbf{Performance Metrics.} The scores achieved by the different models are summarised in Table~\ref{tab:results}. Among the pre-trained models, \emph{Whisper-Large} achieves the best overall WER and chrF scores. However, the best CERand WEPR scores were achieved by the \emph{XLSR-1B} models fine-tuned on CommonVoice 6.1. This aligns with our expectations, as Whisper models are currently the most powerful ASR models, and we expected them to perform best in terms of WER. However, for our use-case, we are more interested in error preservation, thus, CTC-based models without language models are best for preserving the errors. The fine-tuning step on our dataset consisting of learner data yielded a significant boost in performance. It achieves the best WEPR score, which measures the error retention capability. The most comparable model in terms of number of parameters is the XLSR-53 model trained on adult read-aloud data. In comparison to this model, \emph{ChaLL-300M} achieves an improvement of 8 points in WER and a 12-point improvement in WEPR. It is generally the case that larger models perform better. Thus, the interpretation of the results needs to factor this in. As most models are larger than ours, it becomes evident that fine-tuning on learner data increases the performance on this data in general, and the CTC architecture yields a better out-of-the-box preservation of speaker-errors .

\begin{table*}[t!]
    \centering
    \small
    \resizebox{.85\textwidth}{!}{%
    \begin{tabular}{l|l|l}
  & Utterance  &  Err. Type.   \\
\hline
\textsc{Target}   &  Yeah. Uhm it's -- It \textcolor{red}{have} a \textcolor{orange}{Lampe}. Uhm you can -- & has/have, German\\
\textsc{Chall300M} & e uhm it's it's \textcolor{red}{have} a \textcolor{orange}{lampe} you can &  has/have, German\\
\textsc{Whisper-Large} & it has a lamp & - \\
\hline
\hline
\textsc{Target}  & (...) What you're rather be a (...)- able \textcolor{red}{for} fly or be invisible- invisible? & for/to \\
\textsc{Chall300M} & wuld your reader be be aabble \textcolor{red}{for} fly or be invisible invisible  & for/to \\
\textsc{Whisper-Large} & would your reader be able to fly or be invisible & - \\                              
\hline
\hline
\textsc{Target}  &  Do you have \textcolor{red}{a} enemy? & a/an\\
\textsc{Chall300M} & do you have \textcolor{red}{a} enemey & a/an \\
\textsc{Whisper-Large} &do you have an enemy & - \\
\hline
\hline
\textsc{Target}  & What \textcolor{red}{do} you favourite food?& do/is\\
\textsc{Chall300M} & what \textcolor{red}{do} you favorite food  & do/is\\
\textsc{Whisper-Large} & what's your favorite food & - \\
\hline
\end{tabular}
    }
    \caption{Manually selected examples.}
    \label{tab:selected-weprs}
\end{table*}

\noindent\textbf{WEPR Analysis.} To show in more detail the reduction in WEPR, we compare the handling of specific speaker errors. Table~\ref{tab:confusion_ratio} shows the confusion for the 20 most frequent examples, that is, the cases where the ASR system corrects a error it should have preserved. For each type of confusion, we report the rate at which it occurs. For instance, when the speaker mistakenly said "have" (denoted "have@!"), \emph{ChaLL-300M} corrected it to "has" in 1.5\% of cases, Whisper-Large corrected it in 23.1\% of cases, and XLSR-1B in 12.9\% of cases. Thus, \emph{ChaLL-300M} preserved this particular kind of error the best. In total, it mistakenly corrected 15\% of the 20 most frequent speaker-errors, while \emph{Whisper-Large} corrected 26\%, and \emph{XLSR-1B} corrected 22.9\%. It is interesting to note that two out of total three cases where \emph{XLSR-1B} has the lowest rate of mis-correction is for pronunciation errors ("de@!" and "dis@!"). We also note that a majority of the most frequent unwanted error-corrections are deletions. 

On the other hand, Table~\ref{tab:correct_ratio} shows the frequency at which the ASR systems correctly preserved the errors made by the speakers. For instance, when the speaker mistakenly says "have" (denoted as "have@!"), then  \emph{ChaLL-300M} preserves this error in 87.5\% of cases, while \emph{Whisper-Large} preserves it in only 58.7\% and \emph{XLSR-1B} in only 69.9\% of cases. In total, \emph{ChaLL-300M} is able to preserve 82.7\% of the of the 20 most frequent errors made by speakers, while \emph{Whisper-Large} only preserved 67.3\% of speaker errors and \emph{XLSR-1B} preserved 72.3\%. 

Thus, \emph{ChaLL-300M} displays a strong ability to preserve the errors made by speakers, which is crucial for the downstream task of providing automated corrective feedback.

\subsection{Qualitative Results}\label{subsec:qual-results}

Table~\ref{tab:selected-weprs} shows four manually selected examples, highlighting some errors which the best-performing pre-trained model, \emph{Whisper-Large}, corrects, and our model preserves. In the first example, it shows the error of using "have" instead of "has", as well as using the German pronunciation of the word "lamp" (i.e., "Lampe"). \emph{Whisper-Large} corrects these errors, and creates a grammatically correct English utterance. The \emph{ChaLL-300M} model preserves these errors as desired. The second error is a prepositional error, where the learner said "for fly" instead of "to fly". The \emph{Chall-300M} model correctly preserved this error, while the language model used in \emph{Whisper-Large} smoothed out the error. The third example is an error of the indefinite article: the learner used "a" instead of "an", which \emph{ChaLL-300M} correctly preserved while \emph{Whisper-Large} corrected the error. The final example contains the usage of the wrong verb "do" instead of "is", which again is correctly preserved by our model while Whisper corrects the error.


\section{Conclusion and Outlook}
Our work shows that state-of-the-art ASR systems have difficulties handling young learners' speech; furthermore, they tend to correct the errors made by the speakers, which makes the downstream identification of speaker errors and provision of corrective feedback impossible. Thus, we collected around 85 hours of children's language learner speech data, which we used to fine-tune a custom model. Our model outperforms all the others (including Whisper-Large) in terms of error preservation (Word-Based Error Preservation Rate, WEPR) and surpasses the English models of comparable size ($\approx 300M$ parameters) by a large margin in terms of Word Error Rate. Thus, our research shows the necessity of using targeted data (in this case, children who learn a foreign language) to fine-tune an ASR module, which is useful in downstream tasks. The focus of this work lies in a) investigating the utility of existing systems and b) creating an adequate ASR system that can be used as part of a language learning support tool to increase the students' speaking opportunities. 
As a next step, we will investigate how to enhance error preservation. For this, training larger models is the most straightforward approach. However, we also plan to train the ASR system jointly with error annotations. For this, we started the creation of more detailed error annotations. Initial results have shown that verbal errors are the largest error category for young learners of English in Switzerland (with about 22\% of all errors) , and within these, wrong subject-verb agreement is most frequent.
Similarly, investigating how to handle frequent code-switching to German words or sentence fragments is an unsolved issue that needs to be addressed to improve downstream tasks. Even \emph{Whisper-Large}, which can handle multiple languages in principle, did not perform well in detecting code-switching.

Finally, we aim to evaluate  ASR models in the context of integrating them with a conversational agent and corrective feedback. 

\section*{Acknowledgments}
This work was supported by the Swiss Innovation Agency (Innosuisse) within the project "Towards a Voice-Based Chatbot for Language Learners (ChaLL)" [102.580.1 IP-ICT]. We thank the teachers and students who were part of the data collection for their efforts.

\section*{Limitations}
While offering a unique tool for error-preserving ASR of young language learners, this work presents itself with a few limitations. 

\paragraph{Limited Demographic.} The dataset stems from a specific demographic of Swiss school children learning English in grades 4 to 6. An extension of the work would include language learners from other countries/with an academic language other than German/with a different language of instruction, or a larger range of ages. Thus, the transferability of our results must be confirmed with a different dataset.

\paragraph{Outsourcing Error Annotation.} The outsourcing of transcription and error annotations always poses a risk of yielding erroneous data, since most transcribers are not trained in error annotation. We mitigated this risk by providing comprehensive guidelines and a steady exchange with the transcription agency. However, we plan to enhance the error annotations with a more detailed label set and annotators trained in this task.

\paragraph{Small Model.} Due to the high computational cost of fine-tuning a 1B parameter model, we limited ourselves to fine-tuning the 300M parameter XLSR model. Most research indicates that the usage of larger models yields better results; thus, there is still potential in terms of increasing WER and WEPR. However, our results showed that even a small model can preserve errors better than state-of-the-art pre-trained models, which was the main scope of this work.

\paragraph{No Performance Tuning.} Since the scope of this work is to understand if the usage of young learners' speech data is beneficial for our purposes, we did not tune the performance of our model. That is, we did not perform any hyper-parameter tuning or any other methods to increase performance (e.g., joint prediction of errors using a language model). Thus, there is still a large margin of improvement using our dataset.

\paragraph{Data Availability.} Since our data consists of children's spontaneous speech, we must ensure its protection. Thus, we cannot make it freely available. While we publicly release the models trained on the data, access to the transcripts and recordings can only be granted in the scope of a joint project, subject to a collaboration agreement.

\section*{Ethical Considerations}
The main risks in this project have to do with data protection: all speakers are minors between 9 and 14 years of age, so their personal data must be very well safeguarded. Therefore, key government institutions approved the data collection before speakers were recruited, and informed consent was obtained from each speaker's legal caretaker (cp. details in Section \ref{subsec:audio-recording}). 
Consent forms entailed information about the nature of the project and data collection procedures, as well as a comprehensive description of the legal principles we followed to collect, use, and store voice data, transcripts, and annotations. The data protection measures we implemented for security and confidentiality were fully disclosed (e.g. password-protected documents, pseudonymisation, firewalls etc.) and risks to participants (e.g. potential voice recognition by project members) were outlined. Voice data and transcripts were pseudonymised by those project members who act as data owners before sharing them with other research partners and third parties. 
Third-party access to the collected data will be enabled in a closely controlled setting consisting of a joint project with a collaboration agreement.

\section*{Use of AI Assistants}
ChatGPT was used to support the creation of some figures. No AI assistants were used to write the text of this paper.


\bibliography{custom}



\end{document}